\documentclass{article}

\PassOptionsToPackage{numbers, compress}{natbib}

\usepackage{subcaption}
\usepackage[preprint]{main}

\usepackage[utf8]{inputenc} 
\usepackage[T1]{fontenc}    
\usepackage{hyperref}       
\usepackage{url}            
\usepackage{booktabs}       
\usepackage{amsfonts}       
\usepackage{nicefrac}       
\usepackage{microtype}      
\usepackage{xcolor}         
\usepackage{graphicx} 
\usepackage{multirow}
\usepackage{amssymb}  
\usepackage{amsmath}
\usepackage{multirow}

\definecolor{myblue}{HTML}{8DCCFF}

\title{Rig3R: Rig-Aware Conditioning and Discovery \\
for 3D Reconstruction}

\title{Rig3R: Rig-Aware Conditioning for Learned 3D Reconstruction}

\author{%
  Samuel Li\thanks{Equal contribution, order decided by coin toss. Work done while at Wayve.}$\ ^{\ 1,2}$ 
  \quad 
  Pujith Kachana$^{*1,2}$ 
  \quad 
  Prajwal Chidananda$^1$ 
  \quad 
  Saurabh Nair$^1$ 
  \\ 
  \textbf{Yasutaka Furukawa}$^1$ 
  \quad 
  \textbf{Matthew Brown}$^1$
  \\\\
 $^1$ Wayve Technologies \quad $^2$ Carnegie Mellon University
}

\begin{document}

\maketitle

\begin{abstract}\vspace{-1em}
Estimating agent pose and 3D scene structure from multi-camera rigs is a central task in embodied AI applications such as autonomous driving. Recent learned approaches such as DUSt3R have shown impressive performance in multiview settings. However, these models treat images as unstructured collections, limiting effectiveness in scenarios where frames are captured from synchronized rigs with known or inferable structure.
To this end, we introduce \textbf{Rig3R}, a generalization of prior multiview reconstruction models that incorporates rig structure when available, and learns to infer it when not. Rig3R conditions on optional rig metadata including camera ID, time, and rig poses to develop a rig-aware latent space that remains robust to missing information. It jointly predicts pointmaps and two types of raymaps: a pose raymap relative to a global frame, and a rig raymap relative to a rig-centric frame consistent across time. Rig raymaps allow the model to infer rig structure directly from input images when metadata is missing.
Rig3R achieves state-of-the-art performance in 3D reconstruction, camera pose estimation, and rig discovery—outperforming both traditional and learned methods by  17--45\% mAA across diverse real-world rig datasets, all in a single forward pass without post-processing or iterative refinement. 

\end{abstract}
    
\begin{figure*}[ht]
  \centering
   \includegraphics[width=1\linewidth]{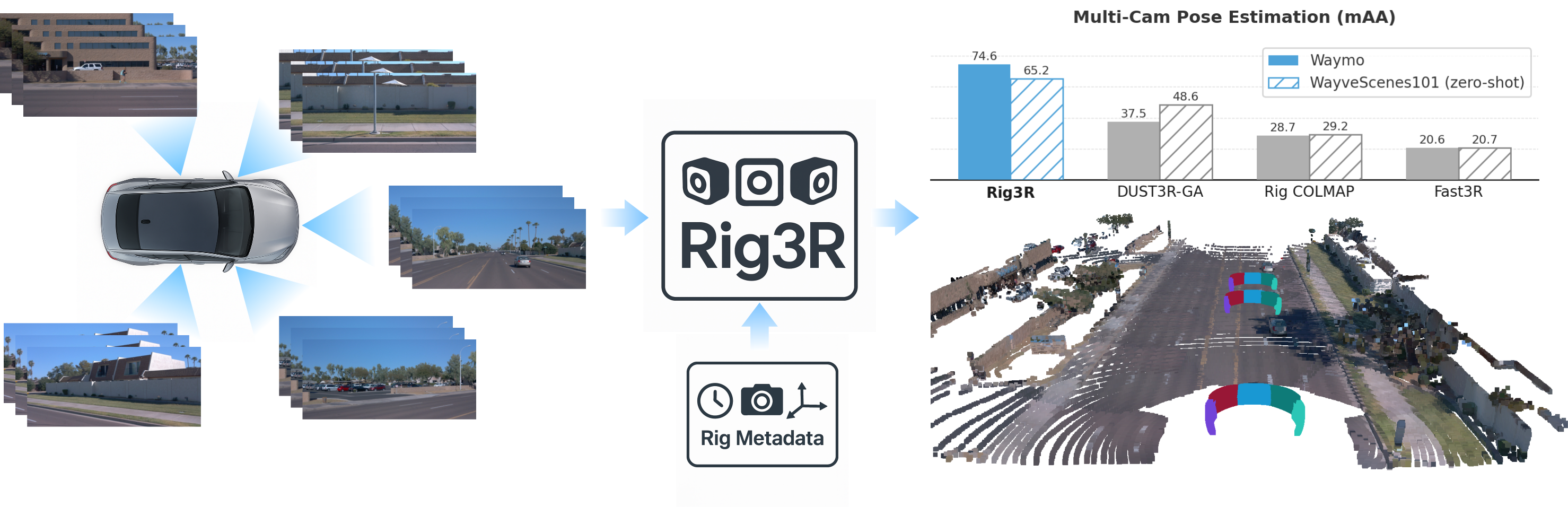}
    \caption{
    Rig3R is the first learned 3D vision model to leverage rig constraints when available, and the first method to support rig calibration discovery from unordered images when they are not—achieving strong 3D consistency and performance across diverse settings and rig configurations.
    }
   \label{fig:hero_fig}
\end{figure*}

\section{Introduction}
\label{sec:intro}

Multi-view scene estimation of camera poses and 3D structure from images is a core capability in computer vision and has enabled spatial understanding for embodied agents, robotic systems, and large-scale visual localization~\cite{hartley2003multiple}. Accurate estimation of structure and motion is essential for tasks such as simultaneous localization and mapping (SLAM)\cite{orb-slam2, loam}, scene relocalization and view synthesis applications \cite{nerf, gsplat, instant-ngp, neus}. Traditional pipelines based on Structure-from-Motion (SfM) and Multi-View Stereo (MVS) reconstruct scenes by optimizing for geometry via matched image features~\cite{Schonberger_2016_CVPR, MurArtal2015ORB, furukawa2010}. While effective in controlled settings, these methods are brittle in the presence of dynamic objects, visual repetition, or feature poor environments, and often require careful tuning.

Recent learned methods such as DUSt3R~\cite{dust3r_cvpr24} have shown impressive capabilities in multiview 3D reconstruction, with many successors~\cite{yang2025fast3r, tang2024mv} extending this to single-pass inference.
A limitation, however, is that these works treat images as unstructured collections.
This overlooks a key structural prior common in real-world applications: images are often captured using synchronized multi-camera rigs with fixed relative configurations. Rig metadata—such as camera ID, timestamp, and relative poses—can provide valuable cues, especially when field-of-view overlap is limited. While classical pipelines can exploit this structure~\cite{kaess2010probabilistic, heng2015leveraging, carrera2011slam}, feedforward models currently leave it untapped.

We introduce \textbf{Rig3R}, a transformer-based model for multiview 3D reconstruction and pose estimation that leverages rig metadata when available and learns to infer rig structure when it is not. Rig3R handles unstructured image sets, calibrated rigs, and everything in between, predicting dense pointmaps and raymaps for each image in a single forward pass. These raymaps spatially encode camera intrinsics and extrinsics, which can be recovered in closed form—even in ambiguous regions such as sky or dynamic pixels. To enable this flexibility, Rig3R combines metadata embeddings with dropout training, and includes a dedicated rig prediction head that infers rig structure directly from image content when metadata is unavailable.

Our key contributions are:
\begin{itemize}


\item The first learned method that leverages rig constraints to improve 3D reconstruction and pose estimation, while generalizing to inputs with partial or missing metadata (e.g., camera ID, timestamp, rig poses).

\item A novel output representation based on global and rig-relative raymaps, enabling closed-form pose estimation and rig structure discovery from unordered image inputs.

    \item Extensive experiments across diverse real-world driving datasets show that Rig3R achieves state-of-the-art performance in 3D reconstruction, camera pose estimation, and rig discovery, outperforming both traditional and learned methods, all in a single forward pass.
\end{itemize}

\section{Related Works}
\label{sec:related-works}







\textbf{Multi-View 3D Reconstruction.}  
Classical pipelines follow a two-stage paradigm: Structure-from-Motion (SfM) for sparse pose and point recovery, followed by Multi-View Stereo (MVS) for densification. Systems like COLMAP~\cite{Schonberger_2016_CVPR, glomap, fastmap} rely on feature matching, triangulation, and bundle adjustment, but remain sensitive to occlusion, motion, and low-texture regions~\cite{hartley2003multiple}. Early learning-based methods improved robustness by introducing learned features and matching~\cite{superpoint, superglue, pixloc, loftr, roma}.  
Photometric representations such as NeRFs~\cite{nerf, nerf-wild, das3r} and Gaussian splats~\cite{gsplat, mv-splat} reconstruct scenes via view synthesis but typically require accurate camera poses. More recent approaches bypass both explicit feature matching and known poses, predicting 3D structure directly from RGB images~\cite{wang2024moge, oquab2023dinov2, yang2024depthanything, wang2024vggsfm}. DUSt3R~\cite{dust3r_cvpr24} pioneered pointmap regression from single image pairs without known poses, with follow-up works addressing multi-frame input~\cite{cut3r, spann3r, monst3r}, dynamic scenes~\cite{monst3r, stereo4d, easi3r}, and downstream tasks~\cite{slam3r, mast3r, mast3r-slam}.  
MV-DUSt3R~\cite{tang2024mv} introduces multi-frame attention, while Fast3R~\cite{yang2025fast3r} scales to hundreds of views with global consistency. VGGT~\cite{wang2025vggt} jointly predicts depth, pose, and structure using a transformer backbone. Pow3R~\cite{pow3r} improves flexibility through lightweight conditioning on inputs such as intrinsics, relative pose, or depth. While these models support efficient scene understanding, they treat input views as unordered. Rig3R builds on this single-pass design, conditioning on rig metadata and enabling structure discovery even in the absence of such priors.

\textbf{Camera Pose Estimation.}
Traditional pose estimation relies on geometric solvers such as PnP with RANSAC~\cite{lepetit2009epnp, martin1981pnp} and global optimization via bundle adjustment~\cite{MurArtal2015ORB}, but performance is brittle under occlusion, dynamic motion, or sparse correspondences. Learned methods such as PoseNet~\cite{kendall2016posenet} regress 6-DoF poses directly from images \cite{ray-diffusion, reloc3r, mickey}, while unsupervised approaches~\cite{zhou2017unsupervised} optimize photometric losses to jointly estimate depth and ego-motion. Systems like DROID-SLAM~\cite{teed2022droidslamdeepvisualslam} combine differentiable updates with learned features for increased robustness, and several related method adopt a learned SLAM approach \cite{point-slam, nice-slam, splatam}. Several 3D reconstruction models, including those discussed above, also infer poses alongside or through 3D structure~\cite{dust3r_cvpr24, leroy2024mast3r, wang2025vggt, yang2025fast3r, wang2024vggsfm, tang2024mv, ace-zero}. Rig3R extends this trend by predicting dense raymaps that encode per-pixel directions and camera centers, enabling closed-form recovery of intrinsics and extrinsics while enforcing multiview consistency.

\textbf{Rig-Aware Multi-View Geometry.}
Rig constraints can provide strong geometric cues for multi-view reconstruction, enabling more accurate pose estimation and improved robustness in low-overlap or ambiguous settings. Classical works have leveraged such constraints in various ways. COLMAP~\cite{Schonberger_2016_CVPR} incorporates rig structure by modeling the rig as a single moving entity, jointly optimizing global poses while keeping intra-rig calibration fixed through bundle adjustment. Kaess and Dellaert~\cite{kaess2010probabilistic} introduced a probabilistic SLAM framework for multi-camera rigs that models cross-camera feature associations under motion. Carrera et al.~\cite{carrera2011slam} proposed a SLAM-based method for fully automatic extrinsic calibration of multi-camera systems, even with non-overlapping fields of view. Heng et al.~\cite{heng2015leveraging} developed an infrastructure-based calibration method using image-based localization and prebuilt maps, requiring no manual intervention. Rig3R takes a markedly different approach from these prior works, providing rig information as optional embeddings on the input, enabling accurate and generalizable 3D reconstruction across both structured and unstructured multi-camera configurations.

\begin{figure*}[t]
  \centering
  
   \includegraphics[width=1\linewidth]{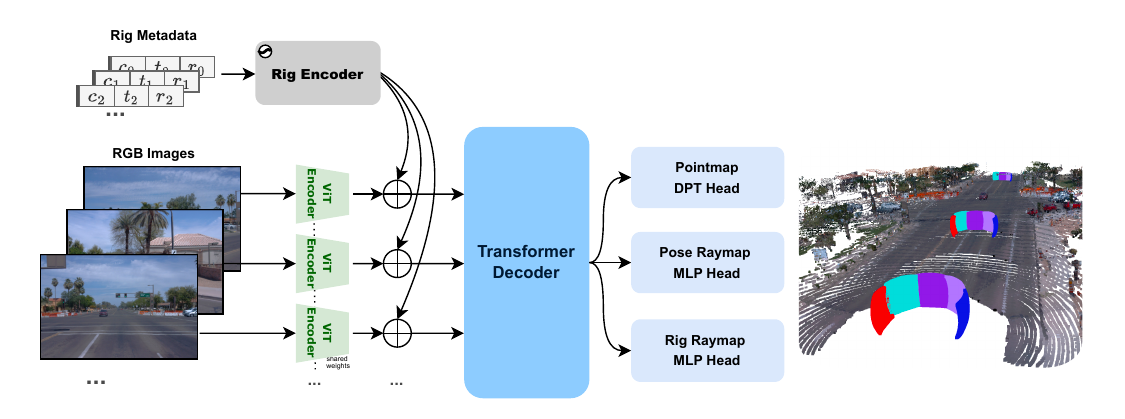}
    \caption{
    Rig3R jointly predicts pointmaps, global raymaps, and rig-relative raymaps, with dropout conditioning on rig embeddings. Global raymaps are color-coded by the discovered rig structure.
    }
   \label{fig:model}
\end{figure*}

\section{Rig-aware 3D Reconstruction}
\label{sec:method}


We address the task of predicting 3D structure and camera poses from a set of \( N \) RGB images \( \{I_i\}_{i=1}^N \), where \( I_i \in \mathbb{R}^{3 \times H \times W} \). Inputs may range from unordered image collections to temporally distributed views captured by multi-camera rigs. Each image may optionally include metadata \( M_i = \{c_i, t_i, r_i\} \), where \( c_i \) is a camera ID, \( t_i \) is a timestamp, and \( r_i \) is a rig-relative raymap encoding the camera pose. Each metadata field is optional and may be independently omitted during training and inference.

Given inputs \( \{I_i, M_i\}_{i=1}^N \), the model predicts for each image: a pointmap \( P_i \in \mathbb{R}^{3 \times H \times W} \) representing per-pixel 3D coordinates in the first image's frame; a confidence map \( C_i \in \mathbb{R}^{H \times W} \) used to weight the pointmap loss; a pose raymap \( R_i^{\text{pose}} \in \mathbb{R}^{H \times W \times 6} \) encoding camera parameters relative to the first image's frame; and a rig raymap \( R_i^{\text{rig}} \in \mathbb{R}^{H \times W \times 6} \) encoding camera parameters relative to a rig-centric frame, decoupled from ego-motion.
Together, these outputs form Rig3R’s predictions:
\[
\text{Rig3R} : \{I_i, M_i\}_{i=1}^N \rightarrow \{P_i, C_i, R_i^{\text{pose}}, R_i^{\text{rig}}\}_{i=1}^N
\]
The following sections detail our raymap representation, metadata encoding, and model architecture.



\subsection{Raymap Representation}
\label{sec:raymaps}


Rig3R outputs dense raymaps from both its pose and rig heads. A raymap is a directional field that assigns a unit ray direction to each pixel, with all rays originating from a shared camera center. This representation encodes both camera intrinsics and pose in a unified, geometrically consistent format.

For each pixel \( (u, v) \), the viewing ray \( \hat{\mathbf{r}}_{uv} \in \mathbb{S}^2 \) is computed as \( \hat{\mathbf{r}}_{uv} = \mathbf{R} \cdot \mathbf{K}^{-1} [u, v, 1]^\top \), where \( \mathbf{K} \in \mathbb{R}^{3 \times 3} \) is the intrinsic matrix, \( \mathbf{R} \in \mathrm{SO}(3) \) is the rotation matrix, and the output is unit-normalized. All rays share a common camera center \( \mathbf{c} \in \mathbb{R}^3 \).

Raymaps offer key advantages over alternative representations. They provide spatially aligned, per-pixel supervision and serve as a stable signal even in ambiguous regions such as sky or dynamic objects. Unlike pointmaps, which infer pose indirectly through 3D predictions and often fail in such regions, raymaps offer a more direct and consistent representation for pose estimation. They also encode interpretable geometry, enabling closed-form recovery of camera intrinsics and extrinsics from ray directions and pixel distances. In our implementation, we recover focal lengths using angular constraints derived from pixel–ray correspondences, and estimate rotations in closed form using SVD from aligned camera and world rays~\cite{arun1987least}. See Section \ref{sec:pose_from_rmap} of supplementary material for more details. 



\subsection{Rig-Aware Metadata}
\label{sec:rig-aware-metadata}

Each image may optionally be associated with a metadata tuple \( M_i = \{c_i, t_i, r_i\} \), where \( c_i \) is a discrete camera identifier shared by all images from the same physical camera, \( t_i \in \mathbb{R} \) is a continuous timestamp normalized in seconds, and \( r_i \in \mathbb{R}^{H \times W \times 6} \) is a rig-relative raymap encoding the camera's pose within the rig.
This metadata provides geometric and temporal context for multiview reasoning. The combination of camera ID and timestamp forms a structured decomposition of frame identity, offering strong cues for spatiotemporal alignment. All metadata fields are optional and can be independently dropped during training to encourage robustness to missing information.

\subsection{Model Architecture}
\label{sec:model-architecture}
\textbf{Image Encoder.} Rig3R (see Fig.~\ref{fig:model}) employs a shared ViT-Large encoder~\cite{vit, vaswani2023attentionneed} to independently patchify and encode each input image using 2D sine-cosine positional encodings. We initialize from DUSt3R~\cite{dust3r_cvpr24}, though other works~\cite{wang2025vggt, tang2024mv, yang2025fast3r} indicate that performance is not sensitive to this choice.

\textbf{Metadata Embedding.} Each patch is optionally augmented with rig-aware metadata: (1) frame index \( N \), (2) camera ID \( c_i \), (3) timestamp \( t_i \), and (4) rig raymap patch \( r_i \). The discrete IDs \( N \) and \( c_i \) are randomly sampled from a larger index range and encoded using 1D sine-cosine embeddings, following~\cite{yang2025fast3r}, enabling generalization to varying numbers of frames and cameras. The timestamp \( t_i \in \mathbb{R} \) is normalized in seconds and encoded similarly. The rig raymap patch \( r_i \in \mathbb{R}^6 \) is linearly projected to the model dimension. All components are concatenated and added to the patch tokens. During training, \( c_i \), \( t_i \), and \( r_i \) are randomly dropped out to promote robustness, while the frame index \( N \) is always included to uniquely identify each image within the transformer.

\textbf{Transformer Decoder.} Patch tokens from all images are passed to a second ViT-Large transformer, trained from scratch, that performs joint self-attention across the full set. This enables Rig3R to aggregate information across views and time, conditioned on metadata when available. Unlike the shared image encoder, the decoder fuses multiview features in a shared latent space.

\textbf{Prediction Heads.} Rig3R uses three multitask heads: one for pointmap prediction and two for raymaps (pose-relative and rig-relative), with shared weights across frames. The pointmap head is a DPT module~\cite{ranftl2021dpt} that predicts a 3D pointmap \( P_i \in \mathbb{R}^{3 \times H \times W} \) and a confidence map \( C_i \in \mathbb{R}^{H \times W} \). Each raymap head consists of two MLPs: one predicts per-pixel ray directions, and the other predicts a global camera center via average pooling over patch tokens. This design avoids dedicated query tokens and ensures all gradients flow through the patch tokens, promoting coherence.

These three outputs are tightly coupled: pointmaps are expected to lie along rays defined by the pose raymap, and rig and pose raymaps are related through ego-motion. This multitask formulation acts as a structural prior, improving consistency and generalization across diverse multiview settings.





\subsection{Training}
\label{sec:training}

\textbf{Training Losses.} 
We train with a multitask loss over pointmaps, pose raymaps, and rig raymaps:
\[
\mathcal{L}_\text{total} = \mathcal{L}_\text{pmap} + \lambda_\text{p} \mathcal{L}_\text{p\_rmap} + \lambda_\text{r} \mathcal{L}_\text{r\_rmap},
\]
where \( \lambda_\text{p} \) and \( \lambda_\text{r} \) are weighting terms for the pose and rig raymap losses.

The pointmap loss \( \mathcal{L}_\text{pmap} \), following \cite{dust3r_cvpr24}, is a confidence-weighted regression objective~\cite{kendall2016uncertainty, novotny2018uncertainty} with scale-normalized ground truth. For frame \( v \), depth-normalized pointmap error is:
\[
\mathcal{L}_\text{pmap} = \sum_{i \in \mathcal{D}^v} C_i^v \left\Vert X_i^v - \frac{1}{\bar{z}} \bar{X}_i^v \right\Vert - \alpha \log C_i^v,
\]
where \( X_i^v \) is the predicted 3D point at pixel \( i \), \( \bar{X}_i^v \) is the ground truth, \( C_i^v \) is the predicted confidence, \( \bar{z} \) is the average scene depth used for normalization, and \( \alpha\) is the weight of the regularization term.

The raymap loss \( \mathcal{L}_\text{rmap} \) includes terms for both ray directions and camera centers:
\[
\mathcal{L}_\text{rmap} = \sum_{h,w} \left\Vert \mathbf{r}_{v,h,w} - \bar{\mathbf{r}}_{v,h,w} \right\Vert + \beta \left\Vert \mathbf{c}_v - \frac{1}{\bar{z}} \bar{\mathbf{c}}_v \right\Vert.
\]
Here, \( \mathbf{r}_{v,h,w} \) is the predicted unit ray direction at pixel \( (h, w) \), and \( \mathbf{c}_v \) is the predicted camera center for frame \( v \). The ground-truth ray direction and camera center are denoted by \( \bar{\mathbf{r}}_{v,h,w} \) and \( \bar{\mathbf{c}}_v \), respectively. The average scene depth \( \bar{z} \) is used for scale normalization, and \( \beta \) weights the center loss term.
Following \cite{wang2025vggt}, we find training is more stable when the model learns scale and direction norms directly—especially for camera centers near the origin.

\textbf{Training Data.} We train Rig3R on a diverse data mix: CO3D-v2~\cite{reizenstein21co3d}, BlendedMVS~\cite{yao2020blendedmvs}, Map-free~\cite{arnold2022mapfree}, ScanNet++ v2~\cite{dai2017scannet}, MVImgNet~\cite{yu2023mvimgnet}, PointOdyssey~\cite{zheng2023pointodyssey}, Virtual KITTI2~\cite{cabon2020virtualkitti2}, TartanAir V2~\cite{tartanair2020iros}, PandaSet~\cite{xiao2021pandaset}, KITTI~\cite{geiger2012kitti}, Argoverse2~\cite{wilson2023argoverse2}, nuScenes~\cite{nuscenes2019}, Waymo~\cite{sun2020waymo}, and an internal dataset. We process relevant driving datasets following \cite{chen2024omnire}. These cover a broad range of scene types—including indoor, driving, synthetic, and object-centric—with an emphasis on data from multi-camera rigs.
For COLMAP datasets, we sample images based on covisibility. For others, we use a random stride within a specified range. In rig-based datasets, we subsample the rig cameras per sequence to increase diversity in rig configurations. Where available, the front-facing camera is always included to ensure overlap and reflect common monocular setups. 

  



\textbf{Embedding Dropout.}
To encourage metadata-aware reasoning and improve robustness, we randomly drop each metadata field (camera ID, timestamp, rig pose) with 50\% probability during training. This structured masking teaches the model to leverage metadata when available, and to infer missing context from image content and cross-view relationships when it is not—enabling generalization to diverse input configurations at inference time.

\textbf{Training Details.}
Rig3R is trained on 24-frame samples with a batch size of 128, using 128 H100 GPUs for 250k steps over 5 days. Images are resized to \(512 \times 512\) with padding. We apply data augmentations including random per-frame color jitter, Gaussian blur, and centered aspect-ratio crops to simulate variation in focal length and image shape. During training, input sequences are randomly shuffled to vary the reference frame and promote generalization. We use the AdamW optimizer with a learning rate of 0.0001 and cosine annealing.
\begin{figure}[t]
  \centering
   \includegraphics[width=1\linewidth]{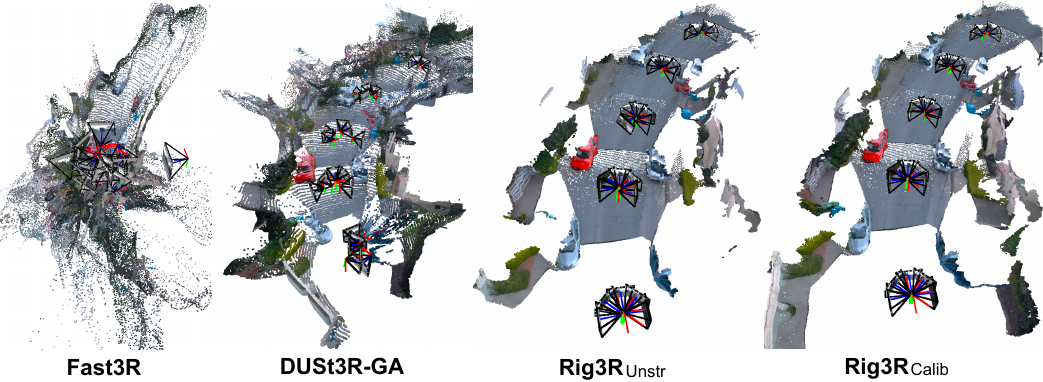}
   \caption{Qualitative results of baselines vs Rig3R with and without rig embeddings. Fast3R fails to find consistent structure and poses. DUSt3R with global alignment gets poses roughly correct, but with inconsistent rig geometry. Rig3R\textsubscript{Unstr} and Rig3R\textsubscript{Calib} show increasingly refined rig geometry.}
   \label{fig:qual-comparison}
\end{figure}


\section{Experiments}
\label{sec:experiments}

\textbf{Evaluation Data.} 
We evaluate Rig3R on the Waymo Open~\cite{sun2020waymo} validation set and WayveScenes101~\cite{zürn2024wayvescenes101datasetbenchmarknovel}, both featuring 5-camera rigs and approximately 200 timesteps per scene at 10 FPS under diverse real-world driving conditions. Waymo provides LiDAR-based ground-truth poses and 3D points, while WayveScenes101 uses COLMAP reconstructions. For each scene, we extract two 24-frame samples, each using the full 5-camera rig spaced approximately 2 seconds apart.

\paragraph{Baselines.}
We compare Rig3R to both learned and classical baselines for multi-view 3D reconstruction and pose estimation. MV-DUSt3R~\cite{tang2024mv} and Fast3R~\cite{yang2025fast3r} are architecturally similar to Rig3R, using transformer-based, feedforward multi-view inference. DUSt3R-GA~\cite{dust3r_cvpr24} predicts stereo pointmaps refined via global optimization. As a classical baseline, we evaluate COLMAP~\cite{Schonberger_2016_CVPR} in both unstructured and rig-aware modes. While learned models operate per sample, we allow COLMAP to process full scenes for stronger global context.
We evaluate two variants of Rig3R: Rig3R\textsubscript{Unstr}, which receives no metadata and treats the sequence as unstructured, and Rig3R\textsubscript{Calib}, which is given full rig metadata (camera ID, timestamp, and rig raymaps). Note that Rig3R\textsubscript{Calib} and Rig COLMAP are the only methods that leverage rig constraints, and both receive the same rig calibration information.

\begin{table*}[t]
    \centering
    \renewcommand{\arraystretch}{1}
    \setlength{\tabcolsep}{4pt}
    \begin{tabular}{l|cccccc|cccccc|c}
        \toprule
        \multirow{3}{*}{Method}
        & \multicolumn{6}{c|}{Waymo}
        & \multicolumn{6}{c|}{WayveScenes101 \textit{(unseen)}} 
        & \multirow{3}{*}{Time} \\
        \cmidrule(lr){2-7} \cmidrule(lr){8-13}
        & \multicolumn{2}{c}{@15° $\uparrow$} & \multicolumn{2}{c}{@5° $\uparrow$} & \multicolumn{2}{c|}{@30° $\uparrow$}
        & \multicolumn{2}{c}{@15° $\uparrow$} & \multicolumn{2}{c}{@5° $\uparrow$} & \multicolumn{2}{c|}{@30° $\uparrow$} \\
        \cmidrule(lr){2-3} \cmidrule(lr){4-5} \cmidrule(lr){6-7}
        \cmidrule(lr){8-9} \cmidrule(lr){10-11} \cmidrule(lr){12-13}
        & RRA & RTA & RRA & RTA & \multicolumn{2}{c|}{mAA} & RRA & RTA & RRA & RTA & \multicolumn{2}{c|}{mAA} & \\
        \midrule
        COLMAP             & 31.1 & 24.4 & 23.0 & 22.5 & \multicolumn{2}{c|}{22.7} & 34.1 & 26.0 & 28.1 & 23.0 & \multicolumn{2}{c|}{24.4} & >2m \\
        MV-DUSt3R          & 44.3 & 23.8 & 18.9 & 8.4 & \multicolumn{2}{c|}{15.8}  & 40.0 & 27.0 & 13.6 & 9.5 & \multicolumn{2}{c|}{13.1} & 10.5s \\
        DUSt3R-GA          & 56.0 & 57.9 & 18.2 & 37.2 & \multicolumn{2}{c|}{37.5} & \textcolor{orange}{89.1} & \textcolor{orange}{61.8} & 33.7 & \textcolor{orange}{47.4} & \multicolumn{2}{c|}{\textcolor{orange}{48.6}} & >2m \\
        Fast3R             & 46.8 & 31.2 & 19.2 & 13.3 & \multicolumn{2}{c|}{20.6} & 61.1 & 29.1 & 23.8 & 12.3 & \multicolumn{2}{c|}{20.7} & 3.9s \\
        Rig3R\textsubscript{Unstr} & \textcolor{orange}{96.6} & \textcolor{orange}{83.9} & \textcolor{orange}{66.0} & \textcolor{orange}{71.6} & \multicolumn{2}{c|}{\textcolor{orange}{74.6}} & 49.2 & 52.4 & 20.5 & 36.0 & \multicolumn{2}{c|}{25.7} & 5.7s \\
        \cmidrule{1-14}
        Rig COLMAP         & 38.6 & 31.1 & 28.4 & 28.6 & \multicolumn{2}{c|}{28.7} & 43.0 & 31.8 & \textcolor{orange}{33.9} & 27.1 & \multicolumn{2}{c|}{29.2} & >2m \\
        Rig3R\textsubscript{Calib} & \textcolor{cyan}{\textbf{99.4}} & \textcolor{cyan}{\textbf{91.6}} & \textcolor{cyan}{\textbf{67.4}} & \textcolor{cyan}{\textbf{77.4}} & \multicolumn{2}{c|}{\textcolor{cyan}{\textbf{82.1}}} & \textcolor{cyan}{\textbf{95.8}} & \textcolor{cyan}{\textbf{75.8}} & \textcolor{cyan}{\textbf{77.7}} & \textcolor{cyan}{\textbf{60.0}} & \multicolumn{2}{c|}{\textcolor{cyan}{\textbf{65.2}}} & 5.7s \\
        \bottomrule
    \end{tabular}
    \caption{
    Multi-view pose estimation results, reporting RRA, RTA, and mAA at various precision levels. Intrinsics are withheld. \textcolor{cyan}{\textbf{Cyan}} and \textcolor{orange}{orange} indicate the best and second-best results, respectively.
    }
    \label{tab:pose_estimation}
\end{table*}

\begin{table*}[t]
    \centering
    \begin{tabular}{l|ccc|ccc}
        \toprule
        \multirow{2}{*}{Method} 
        & \multicolumn{3}{c|}{Waymo} 
        & \multicolumn{3}{c}{WayveScenes101 \textit{(unseen)}} \\
        \cmidrule(lr){2-4} \cmidrule(lr){5-7}
        & Acc. $\downarrow$ & Comp. $\downarrow$ & Chamfer $\downarrow$
        & Acc. $\downarrow$ & Comp. $\downarrow$ & Chamfer $\downarrow$ \\
        \midrule
        MV-DUSt3R           & 1.7 & 24.0 & 12.9 & 6.7 & 38.0 & 19.3 \\
        DUSt3R-GA           & 1.9  & 15.2 & 8.6 & 1.4 & 7.8 & 4.6 \\
        Fast3R              & 1.9    & 5.9 & 3.9 & 0.7    & \textcolor{orange}{5.1} & \textcolor{orange}{2.9} \\
        Rig3R\textsubscript{Unstr}   & \textcolor{orange}{0.2}   & \textcolor{orange}{1.4} & \textcolor{orange}{0.8} & \textcolor{orange}{0.4} & 8.2 & 4.3 \\
        Rig3R\textsubscript{Calib}      & \textcolor{cyan}{\textbf{0.1}}    & \textcolor{cyan}{\textbf{0.2}} & \textcolor{cyan}{\textbf{0.2}} & \textcolor{cyan}{\textbf{0.3}}  & \textcolor{cyan}{\textbf{4.1}} & \textcolor{cyan}{\textbf{2.2}} \\
        \bottomrule
    \end{tabular}
    \caption{
        Multi-view pointmap estimation results. We report accuracy and completeness, and their average as the Chamfer distance. Intrinsics are withheld for all methods.
    }
    \label{tab:pointmap_error}
\end{table*}

\subsection{Camera Pose Estimation}
\label{sec:pose_est}

We evaluate pose estimation using relative rotation and translation accuracy (RRA, RTA) at 15° and 5° thresholds, along with mean average accuracy (mAA) over thresholds up to 30°~\cite{wang2024posediffusion, zhang2022relpose, jin2021image}. These metrics capture both coarse correctness and fine-grained precision. Full results are shown in Table~\ref{tab:pose_estimation}.

On Waymo, Rig3R\textsubscript{Calib} achieves the best performance (82.1 mAA), and maintains high precision even at 5° thresholds. Rig3R\textsubscript{Unstr} ranks second overall, despite lacking rig metadata. Learned baselines exhibit sharp drops at 5°. COLMAP improves with rig constraints, and shows smaller differences between thresholds, consistent with classical optimization's binary convergence behavior. We also report wall-clock time per method to estimate poses. Qualitative results in Fig.~\ref{fig:qual-comparison} show Rig3R’s improved spatial consistency—especially with embeddings—even under large spatial displacement.

We also evaluate on WayveScenes101, an unseen dataset with a novel rig configuration constructed from scenes where COLMAP reconstruction succeeded—potentially favoring classical methods. Despite this, Rig3R\textsubscript{Calib} achieves the best performance across all metrics, followed by DUSt3R-GA, which benefits from global optimization. Rig3R\textsubscript{Unstr} remains competitive, outperforming all feedforward baselines and performing comparably to Rig COLMAP.


These results confirm that Rig3R\textsubscript{Calib} achieves the strongest accuracy and precision across datasets, while Rig3R\textsubscript{Unstr} remains robust without rig metadata. Rig-aware embeddings provide a powerful mechanism for generalization and fine-grained pose estimation in diverse multi-view settings.

\begin{figure*}[t]
  \centering
    \includegraphics[width=1\linewidth]{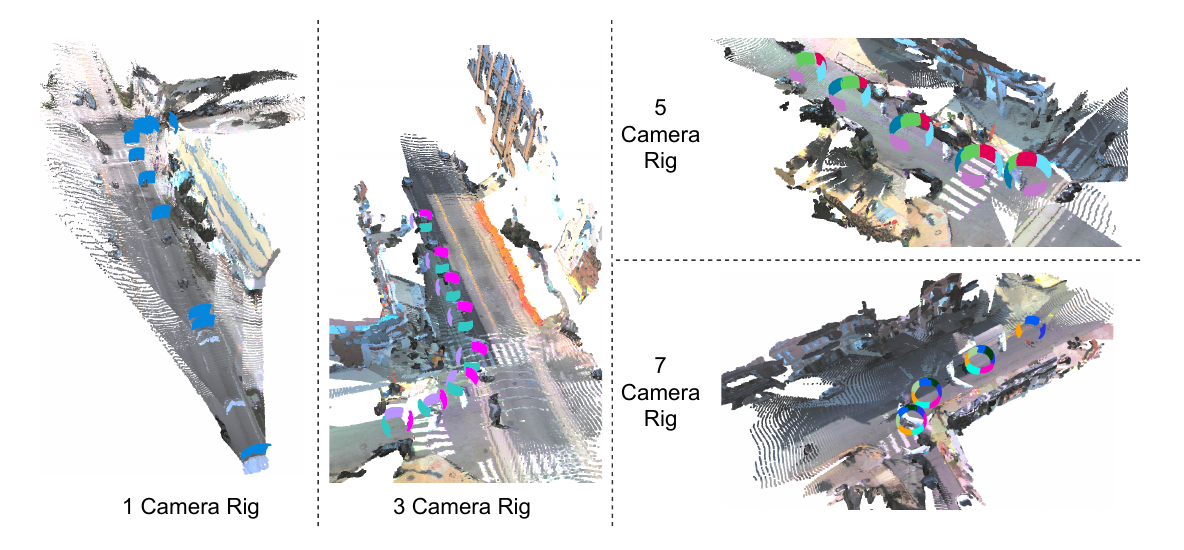}
    \caption{
    Qualitative results from Rig3R across diverse rig configurations, showing pointmaps and global pose raymaps color-coded by camera ID based on clustered rig raymap outputs.
    }
   \label{fig:qual-outputs}
\end{figure*}

\subsection{Pointmap Estimation}
\label{sec:pointmap_estimation}

We evaluate 3D reconstruction quality using pointmap accuracy (Acc.), completeness (Comp.), and Chamfer distance (average of the two). Metrics are computed over sparse pointclouds by masking both predictions and ground-truth to valid regions (see Table~\ref{tab:pointmap_error}).


On Waymo, Rig3R\textsubscript{Calib} achieves the lowest error across all metrics. Rig3R\textsubscript{Unstr} follows closely, significantly outperforming all other baselines and confirming the strength of Rig3R's pointmap predictions even without metadata. Fig.~\ref{fig:qual-comparison} highlights Rig3R’s improvements over baselines in 3D reconstruction quality. In particular, with rig-aware embeddings, the model confidently reconstructs side-view cameras with minimal overlap, where geometric cues alone are often insufficient.

On WayveScenes101, Rig3R\textsubscript{Calib} again leads, demonstrating robust generalization. Interestingly, Fast3R achieves the second-best Chamfer distance, slightly outperforming Rig3R\textsubscript{Unstr} despite lower pose accuracy. This highlights a key advantage of Rig3R: it estimates pose directly from raymaps rather than pointmaps, enabling more consistent multi-view reasoning. See Section \ref{sec:raymap-vs-pointmap} of the supplementary material for further discussion on pose inference from raymaps versus pointmaps.

These results show that Rig3R learns robust scene structure with strong spatial precision and completeness, and that rig metadata further improves reconstruction quality and generalization.



\subsection{Generalization Across Rig Configurations}
\label{sec:rig_generalization}

We assess Rig3R's ability to generalize across rig configurations on the Argoverse2~\cite{wilson2023argoverse2} validation set. We subsample 1, 3, 5, and 7 camera rigs, increasing strides to maintain scene coverage (Fig.~\ref{fig:qual-outputs}).

\begin{figure}[t]
    \centering
    \includegraphics[width=\textwidth]{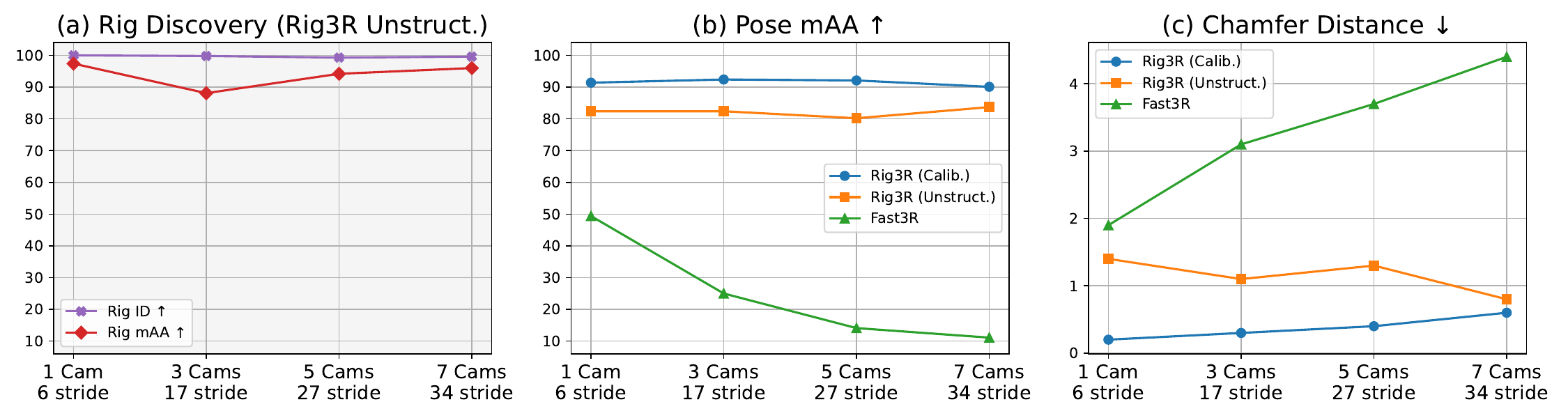}
    \caption{
    Rig3R generalizes across rig configurations. 
    (a) Rig-relative pose mAA and Rig ID clustering accuracy with Rig3R\textsubscript{Unstr}.
    (b) Global pose mAA and (c) Chamfer distance for Rig3R\textsubscript{Calib}, Rig3R\textsubscript{Unstr}, and Fast3R. 
    Both Rig3R variants achieve consistent high performance across all settings.
    }
    \label{fig:rig_discovery_plot}
\end{figure}

\textbf{Rig Calibration Discovery.}
Figure~\ref{fig:rig_discovery_plot}a shows results for Rig3R\textsubscript{Unstr}, which discovers rig calibrations directly from unordered images—without any rig metadata or assumptions about camera configuration. We evaluate two rig-specific metrics: rig ID accuracy and rig-relative pose mAA. Rig ID accuracy reflects how well frames from the same camera are grouped together; we compute it by first clustering the rig raymap outputs, and then evaluating frame assignment accuracy via the Hungarian algorithm. Rig mAA then measures the quality of these predicted clusters, evaluating how accurate the relative orientations and positions of the discovered cameras are in a rig-centeric frame.

Rig3R\textsubscript{Unstr} achieves strong rig mAA and Rig ID accuracy across all configurations. Performance remains strong even as spatial layout becomes more complex, confirming Rig3R's ability to discover diverse rig structures without supervision. To the best of our knowledge, this is the first attempt, learned or classical, to address rig discovery with unordered images and no timestamps. Notably, Rig3R also handles monocular and unordered inputs by predicting identity rig raymaps and producing a single cluster at inference—correctly signaling the absence of a rig. Fig.~\ref{fig:qual-outputs} visualizes various predicted rig configurations, with camera clusters color-coded across time.


\textbf{Flexible Rig Performance.}
We evaluate performance across rig sizes using pose mAA and Chamfer distance (Fig.~\ref{fig:rig_discovery_plot}b–c) for Rig3R\textsubscript{Calib}, Rig3R\textsubscript{Unstr}, and Fast3R. Rig3R\textsubscript{Calib} performs best overall, maintaining high pose and reconstruction quality across all settings. Rig3R\textsubscript{Unstr} also performs strongly, despite receiving no metadata. In contrast, Fast3R degrades as rig size and stride increase—likely due to reduced image overlap, which challenges methods that rely solely on visual correspondence without rig context. These results demonstrate that Rig3R remains robust to spatial variation and generalizes well across diverse rig configurations, as can be seen in Fig.~\ref{fig:qual-outputs}.



\subsection{Ablation Studies}
\label{sec:ablation}

\textbf{Metadata Embeddings.} 
In Table~\ref{tab:embedding_ablation}, we ablate the contribution of each metadata field—camera ID, timestamp, and rig pose—on pose estimation across Waymo and WayveScenes101, which represent previously seen and unseen rig configurations, respectively.

On Waymo, we observe that camera ID and rig pose embeddings provide only modest gains over the unstructured baseline, likely because the rig structure is easily recognized in this familiar setting. In this context, timestamp proves particularly valuable, as it provides dynamic cues for localizing the rig over time. Since the model already reasons about the rig structure implicitly, temporal information becomes the most informative remaining signal.

On WayveScenes101, a previously unseen dataset with a novel rig configuration, we find that camera ID and timestamp embeddings offer only limited gains over the unstructured baseline. This is likely because neither field alone disambiguates the underlying rig layout: camera ID does not indicate motion over time, and timestamp alone does not reveal which camera captured each frame. Classical methods like COLMAP similarly require both to accurately infer rig structure and optimize poses.
In contrast, rig pose embeddings provides a direct spatial signal about the novel rig configuration—crucial for reasoning under domain shift. With this input, Rig3R can recognize and adapt to unseen rig geometries, resulting in a substantial boost in generalization and performance (mAA improves from 25.7 to 56.4). These results highlight the unique role of rig pose metadata in enabling generalization to unseen capture setups.

Providing all metadata yields the best performance across both datasets, confirming that spatial calibration, temporal cues, and view identity are complementary. These results show that Rig3R generalizes well from partial metadata and fully benefits from rig calibration when available.

\textbf{Multi-task Learning.}
We evaluate the impact of each auxiliary head by training three Rig3R variants: with only the pose raymap head (\( \mathcal{L}_\text{pose} \)), with pose + rig raymap (\( \mathcal{L}_\text{pose} + \mathcal{L}_\text{rig} \)), and with pose + pointmap (\( \mathcal{L}_\text{pose} + \mathcal{L}_\text{pmap} \)). As this ablation requires training separate models from scratch, we perform it at reduced scale (batch size 32, 5 datasets) for computational efficiency; results are not directly comparable to full-scale evaluations.
As shown in Table~\ref{tab:head_ablation}, both auxiliary heads individually improve performance in the unstructured setting. The pointmap head provides the largest gain—raising mAA from 45.8 to 53.5—reflecting the value of 3D grounding. The rig head also improves results over using only the pose raymap head, suggesting that even without metadata, it helps maintain a coherent spatial layout. In the calibrated setting, all metrics are higher across the board, indicating some degree of performance saturation. Still, the rig head yields the highest mAA, likely by reinforcing the structure provided by the rig metadata across timesteps, followed by the pointmap head also helps in this setting.
Despite the reduced scale of this setup, we observe a consistent pattern: both heads improve performance, and the rig head is particularly important when rig constraints are available. Importantly, the rig head also enables Rig3R’s novel ability to perform rig discovery. For these reasons, we include both auxiliary heads in the full-scale Rig3R model.

\begin{table*}[t]
    \centering
    \renewcommand{\arraystretch}{1}
    \setlength{\tabcolsep}{3.5pt}
    \begin{tabular}{ccc|cc c c|cc c c}
        \toprule
        \multirow{3}{*}{Cam} & \multirow{3}{*}{Time} & \multirow{3}{*}{Rig}
        & \multicolumn{4}{c|}{Waymo}
        & \multicolumn{4}{c}{WayveScenes101 \textit{(unseen)}} \\
        \cmidrule(lr){4-7} \cmidrule(lr){8-11}
        & & & \multicolumn{2}{c}{@15$^\circ$ $\uparrow$} & \multicolumn{1}{c}{@30$^\circ$ $\uparrow$} & \multicolumn{1}{c|}{$\downarrow$}
              & \multicolumn{2}{c}{@15$^\circ$ $\uparrow$} & \multicolumn{1}{c}{@30$^\circ$ $\uparrow$} & \multicolumn{1}{c}{$\downarrow$} \\
        \cmidrule(lr){4-5} \cmidrule(lr){6-6} \cmidrule(lr){7-7}
        \cmidrule(lr){8-9} \cmidrule(lr){10-10} \cmidrule(lr){11-11}
        & & & RRA & RTA & mAA & Chamfer 
              & RRA & RTA & mAA & Chamfer \\
        \midrule
         &  &              & 96.6 & 83.9 & 74.6 & 0.8 & 49.2 & 52.4 & 25.7 & 4.3  \\
        \checkmark &  &              & 97.0 & 84.3 & 75.1 & 1.1 & 48.0 & 55.4 & 26.8 & 4.5 \\
         & \checkmark &              & 97.6 & \textcolor{cyan}{\textbf{92.7}} & \textcolor{orange}{81.8} & \textcolor{orange}{0.3} & 36.2 & 60.2 & 23.9 & 4.6 \\
         &  & \checkmark & \textcolor{orange}{98.0} & 84.2 & 76.0 & 1.2 & \textcolor{cyan}{\textbf{96.5}} & \textcolor{orange}{66.5} & \textcolor{orange}{56.4} & \textcolor{orange}{2.5} \\
        \checkmark & \checkmark & \checkmark & \textcolor{cyan}{\textbf{99.4}} & \textcolor{orange}{91.6} & \textcolor{cyan}{\textbf{82.1}} & \textcolor{cyan}{\textbf{0.2}} & \textcolor{orange}{95.8} & \textcolor{cyan}{\textbf{75.8}} & \textcolor{cyan}{\textbf{65.2}} & \textcolor{cyan}{\textbf{2.2}} \\
        \bottomrule
    \end{tabular}
    \caption{
    Ablation of input metadata (cam ID, time, rig pose) on pose and pointmap estimation results.
    }
    \label{tab:embedding_ablation}
\end{table*}

\begin{table*}[t]
    \centering
    \renewcommand{\arraystretch}{1.1}
    \setlength{\tabcolsep}{4pt}
    \begin{tabular}{l|ccccc|ccccc}
        \toprule
        \multirow{2}{*}{Variant}
        & \multicolumn{5}{c|}{Unstructured} & \multicolumn{5}{c}{Calibrated} \\
        \cmidrule(lr){2-6} \cmidrule(lr){7-11}
        & \multicolumn{2}{c}{@15$^\circ$ $\uparrow$} & \multicolumn{2}{c}{@5$^\circ$ $\uparrow$} & @30$^\circ$ $\uparrow$
        & \multicolumn{2}{c}{@15$^\circ$ $\uparrow$} & \multicolumn{2}{c}{@5$^\circ$ $\uparrow$} & @30$^\circ$ $\uparrow$ \\
        \cmidrule(lr){2-3} \cmidrule(lr){4-5} \cmidrule(lr){6-6}
        \cmidrule(lr){7-8} \cmidrule(lr){9-10} \cmidrule(lr){11-11}
        & RRA & RTA & RRA & RTA & mAA & RRA & RTA & RRA & RTA & mAA \\
        \midrule
        \( \mathcal{L}_\text{pose} \)                    & 89.2 & 54.1 & 59.6 & 46.7 & 45.8 & \textcolor{cyan}{\textbf{98.6}} & 89.0 & 63.6 & 79.7 & 78.9 \\
        \( \mathcal{L}_\text{pose} + \mathcal{L}_\text{rig} \)  & \textcolor{orange}{90.4} & \textcolor{orange}{56.6} & \textcolor{orange}{60.2} & \textcolor{orange}{49.3} & \textcolor{orange}{48.5} & \textcolor{orange}{98.5} & \textcolor{cyan}{\textbf{91.7}} & \textcolor{orange}{64.8} & \textcolor{cyan}{\textbf{84.1}} & \textcolor{cyan}{\textbf{81.9}} \\
        \( \mathcal{L}_\text{pose} + \mathcal{L}_\text{pmap} \) & \textcolor{cyan}{\textbf{91.3}} & \textcolor{cyan}{\textbf{62.1}} & \textcolor{cyan}{\textbf{61.7}} & \textcolor{cyan}{\textbf{52.5}} & \textcolor{cyan}{\textbf{53.5}} & 98.3 & \textcolor{orange}{90.1} & \textcolor{cyan}{\textbf{66.6}} & \textcolor{orange}{79.8} & \textcolor{orange}{79.8} \\
        \bottomrule
    \end{tabular}
    \caption{
    Model ablation of pointmap (\( \mathcal{L}_\text{pmap} \)) and rig (\( \mathcal{L}_\text{rig} \)) heads on the pose raymap head (\( \mathcal{L}_\text{pose} \)).
    }
    \label{tab:head_ablation}
\end{table*}

\section{Conclusion}
\label{sec:conclusion}

We present Rig3R, a transformer-based model for multiview 3D reconstruction and pose estimation that introduces rig-aware conditioning and rig discovery. Rig3R is the first method to leverage rig metadata in a learned setting and the first to perform rig structure discovery from completely unconstrained image inputs. It jointly predicts pointmaps, global raymaps, and rig-relative raymaps in a single forward pass, achieving strong performance through its spatially-grounded representations.

\textbf{Limitations and Future Work.}
\label{sec:limitations}
Rig3R's main performance limitation is data diversity and quality, particularly regarding variety in rig configurations across existing datasets. One promising direction is to incorporate augmentations that simulate diverse rigs across a continuous configuration space. Future work may also explore balancing structured, rig-based temporal sampling with unordered, general sampling to improve generalization and adaptability across capture settings. While raymaps implicitly downweight dynamic content, explicitly modeling motion could further improve robustness in highly dynamic scenes. Overall, we see rig-aware embeddings as a powerful and generalizable cue, readily applicable to existing and future transformer-based models for multiview reasoning.

\clearpage
\appendix

{\huge\bfseries Appendix\par\vspace{1em}}

The supplementary document provides A) additional visualization of our reconstructions; B) details of the pose estimation algorithm from raymaps; C) discussion on pose inference from raymaps versus pointmaps; and D) an additional experiment evaluating the robustness to calibration errors by injecting Gaussian noise.

\begin{figure*}[ht]
    \centering

    \begin{minipage}[b]{0.49\linewidth}
        \centering
            \includegraphics[width=\linewidth]{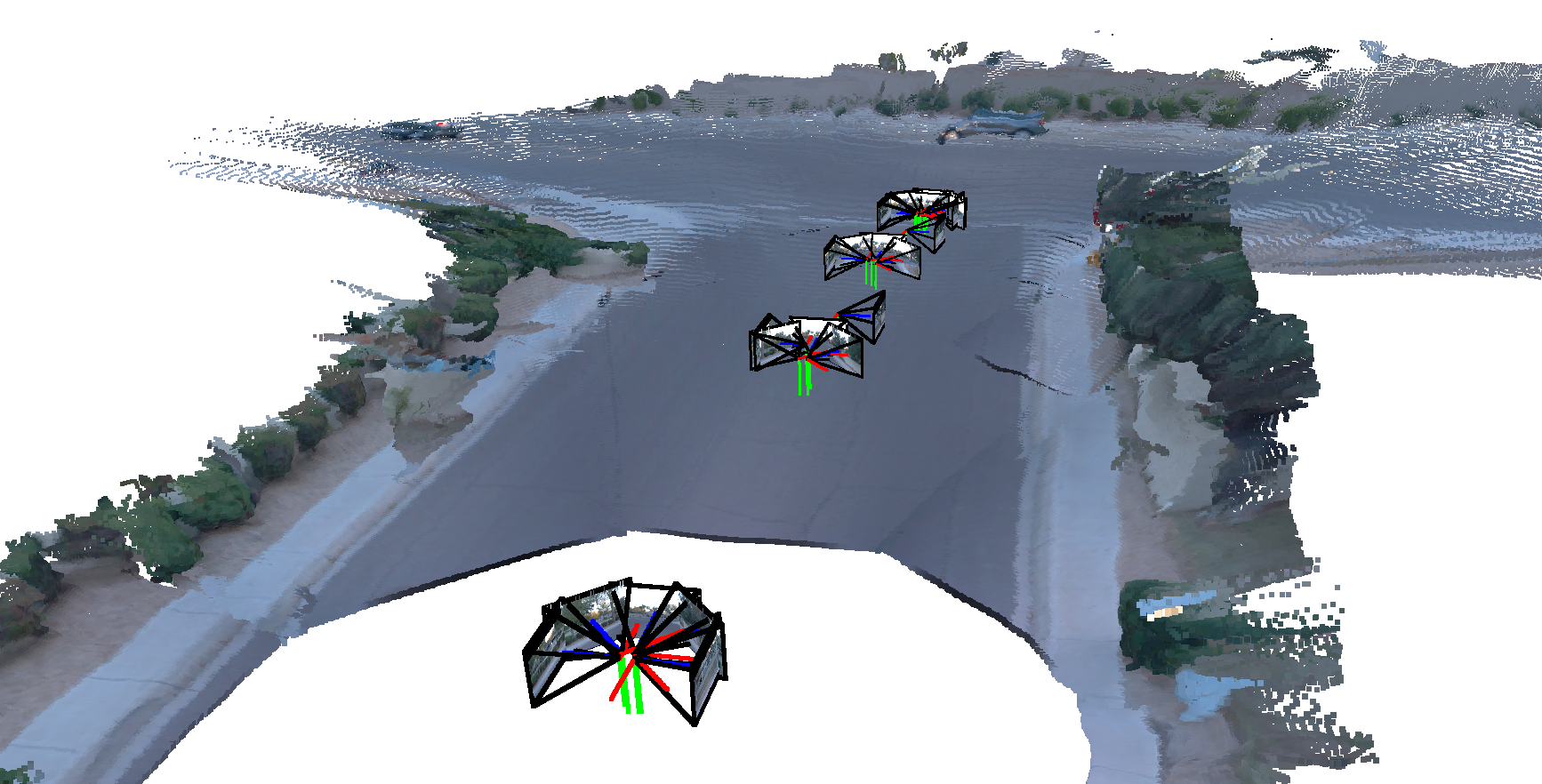}
        \caption*{Unstructured}
    \end{minipage}
    \hfill
    \begin{minipage}[b]{0.49\linewidth}
        \centering
        \includegraphics[width=\linewidth]{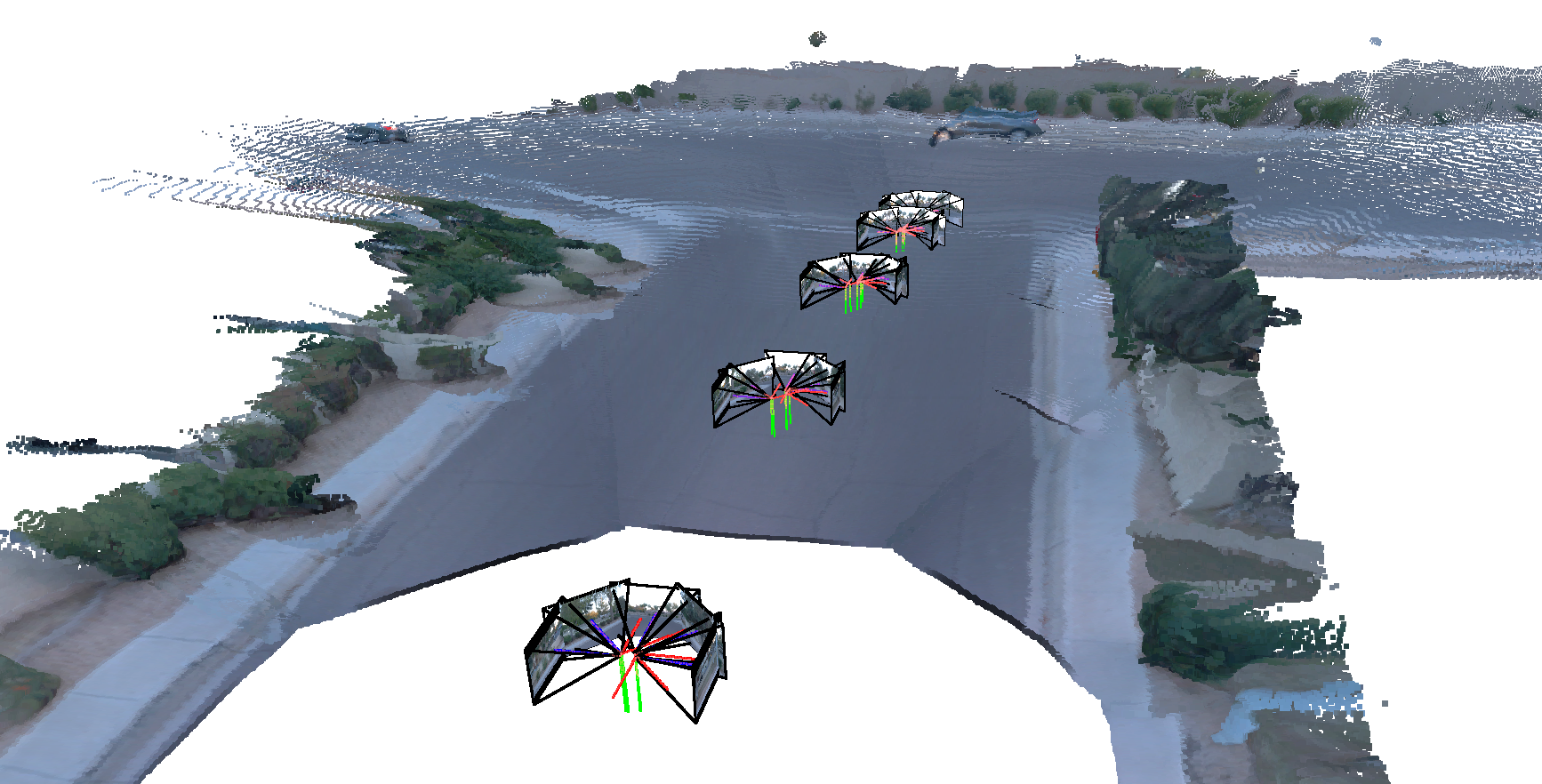}
        \caption*{Calibrated}
    \end{minipage}
    
    \vspace{1em}

    \begin{minipage}[b]{0.32\linewidth}
        \centering
        \includegraphics[width=\linewidth]{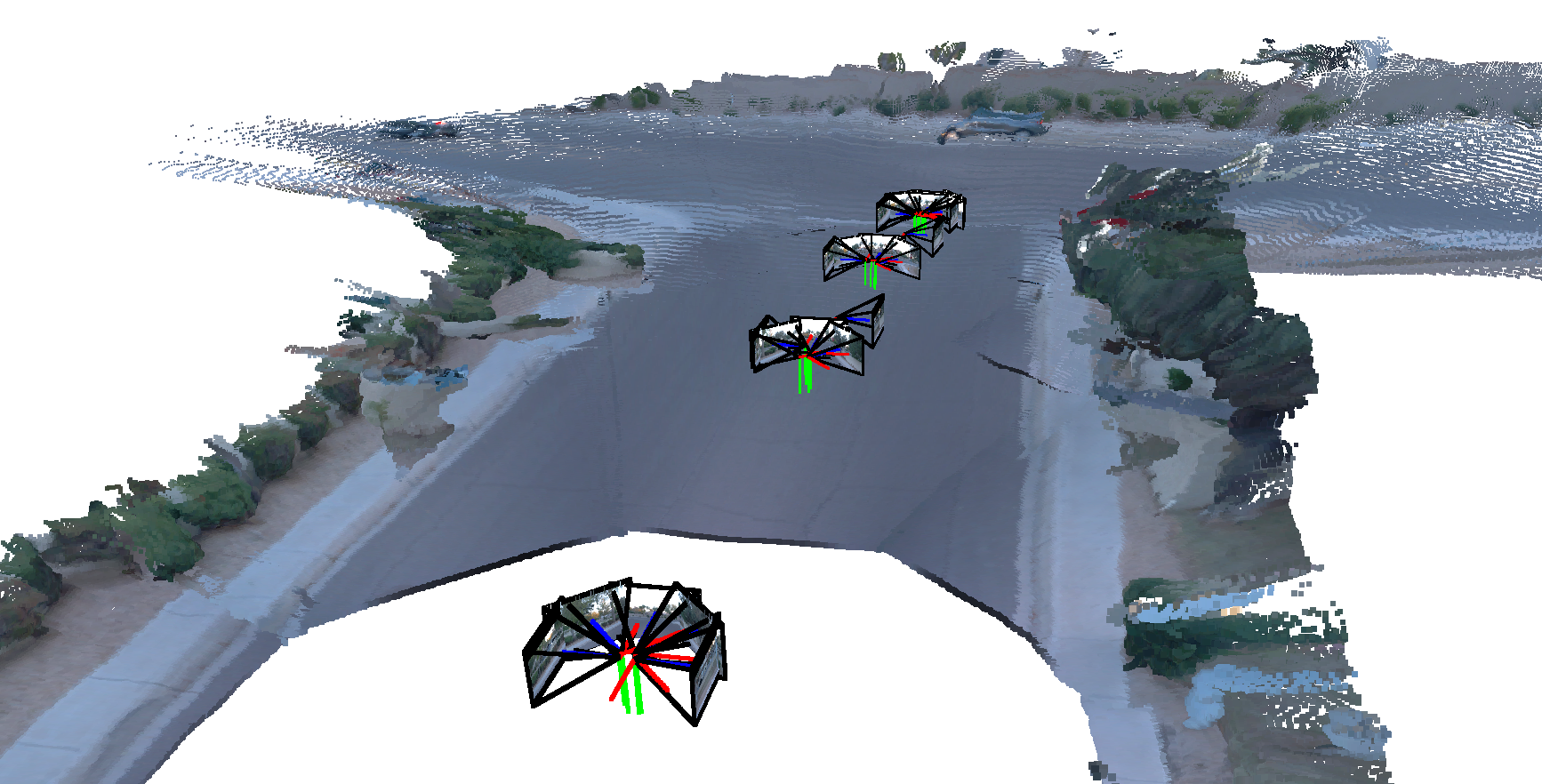}
        \caption*{Cam Embeddings}
    \end{minipage}
    \hfill
    \begin{minipage}[b]{0.32\linewidth}
        \centering
        \includegraphics[width=\linewidth]{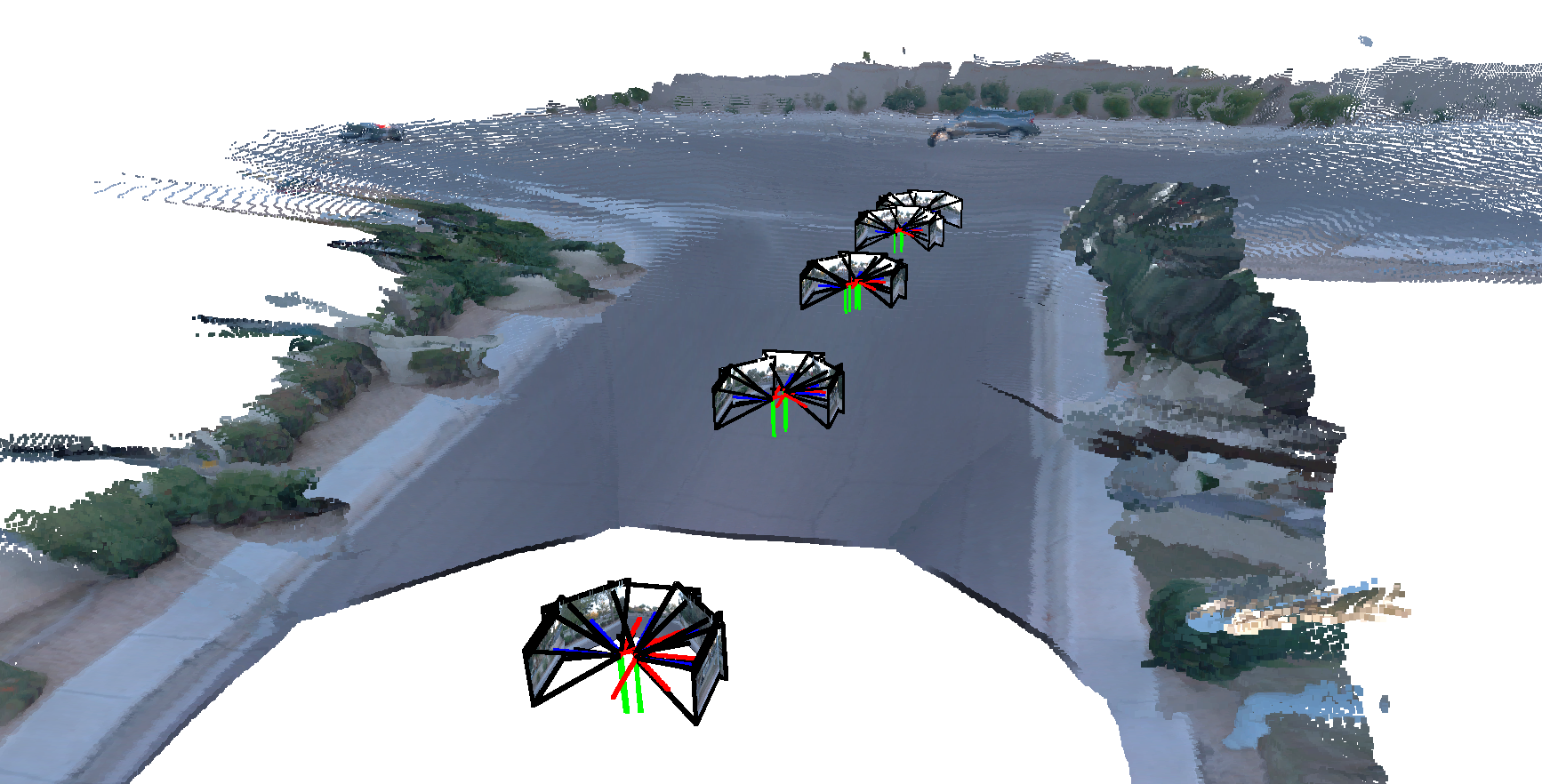}
        \caption*{Time Embeddings}
    \end{minipage}
    \hfill
    \begin{minipage}[b]{0.32\linewidth}
        \centering
        \includegraphics[width=\linewidth]{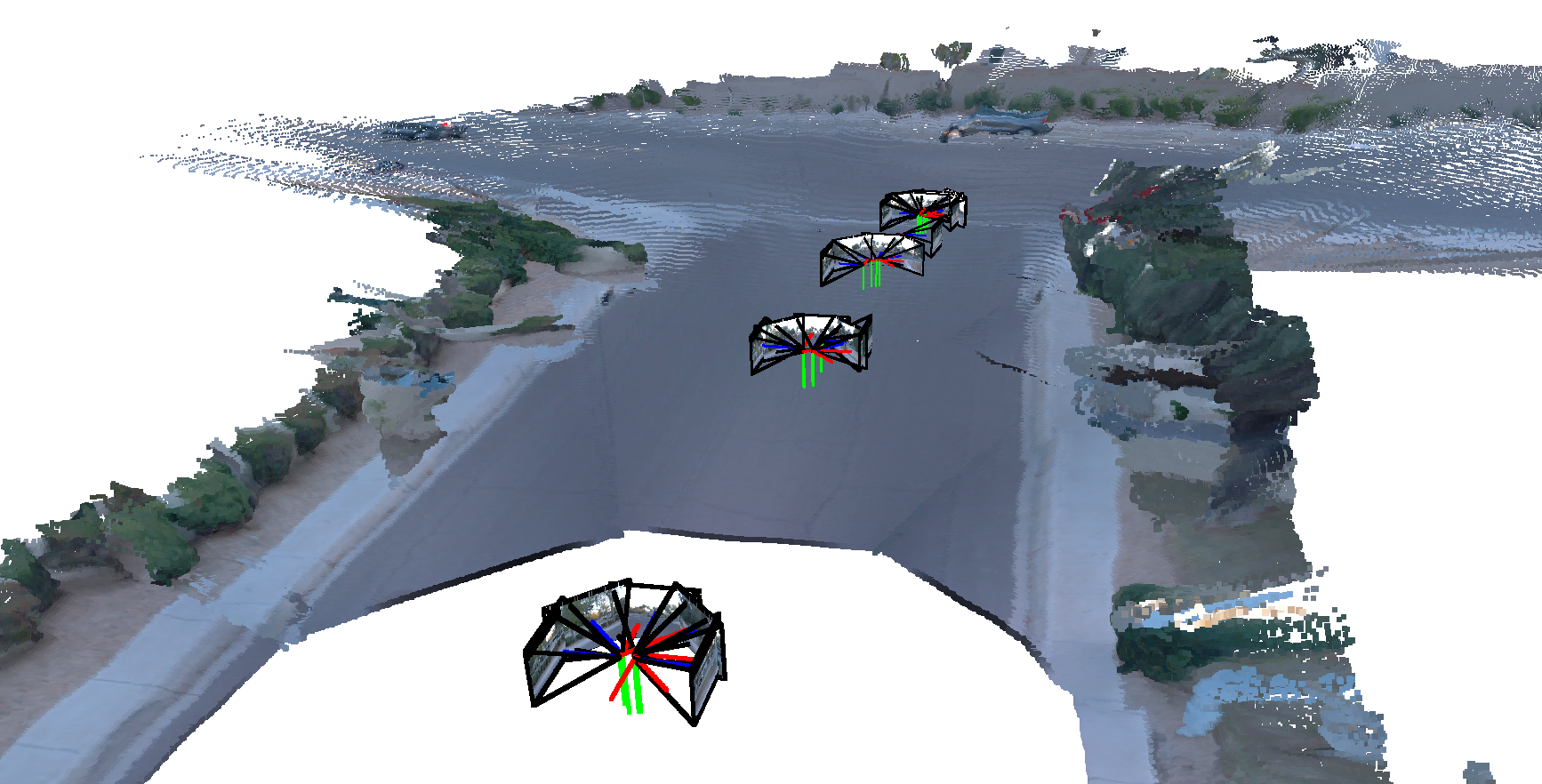}
        \caption*{Rig Embeddings}
    \end{minipage}

    \caption{Visualizations of the qualitative effects of rig metadata embeddings on the Waymo validation set. We observe that with added embeddings, the quality of the estimated poses noticeably improves, and the fine details of reconstructed scene are also better captured.}
    \label{fig:images_embed_ablation}
\end{figure*}

\begin{figure}[ht]
    \centering
    \begin{minipage}[b]{0.32\linewidth}
        \centering
        \includegraphics[width=\linewidth]{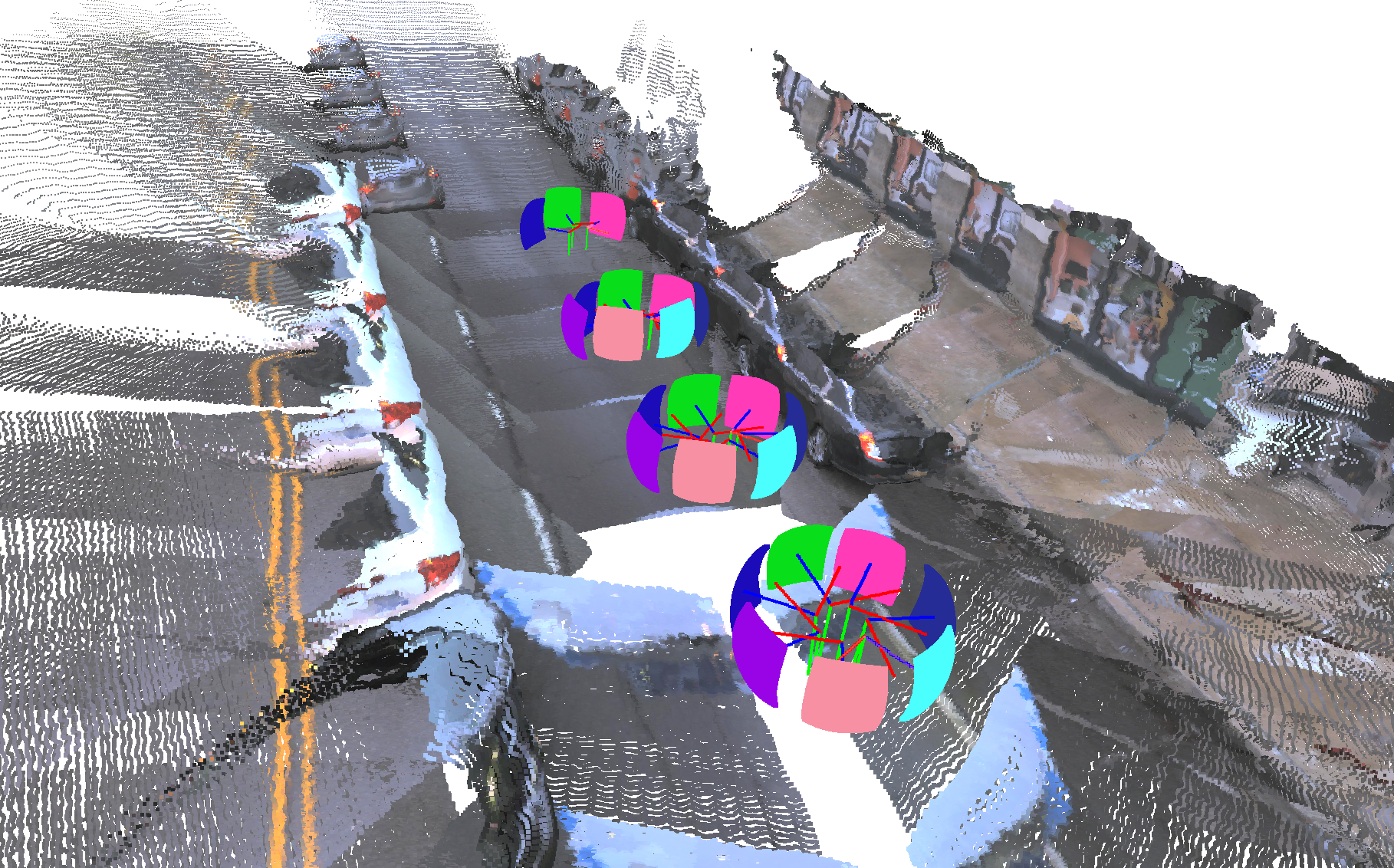}
    \end{minipage}
    \hfill
    \begin{minipage}[b]{0.32\linewidth}
        \centering
        \includegraphics[width=\linewidth]{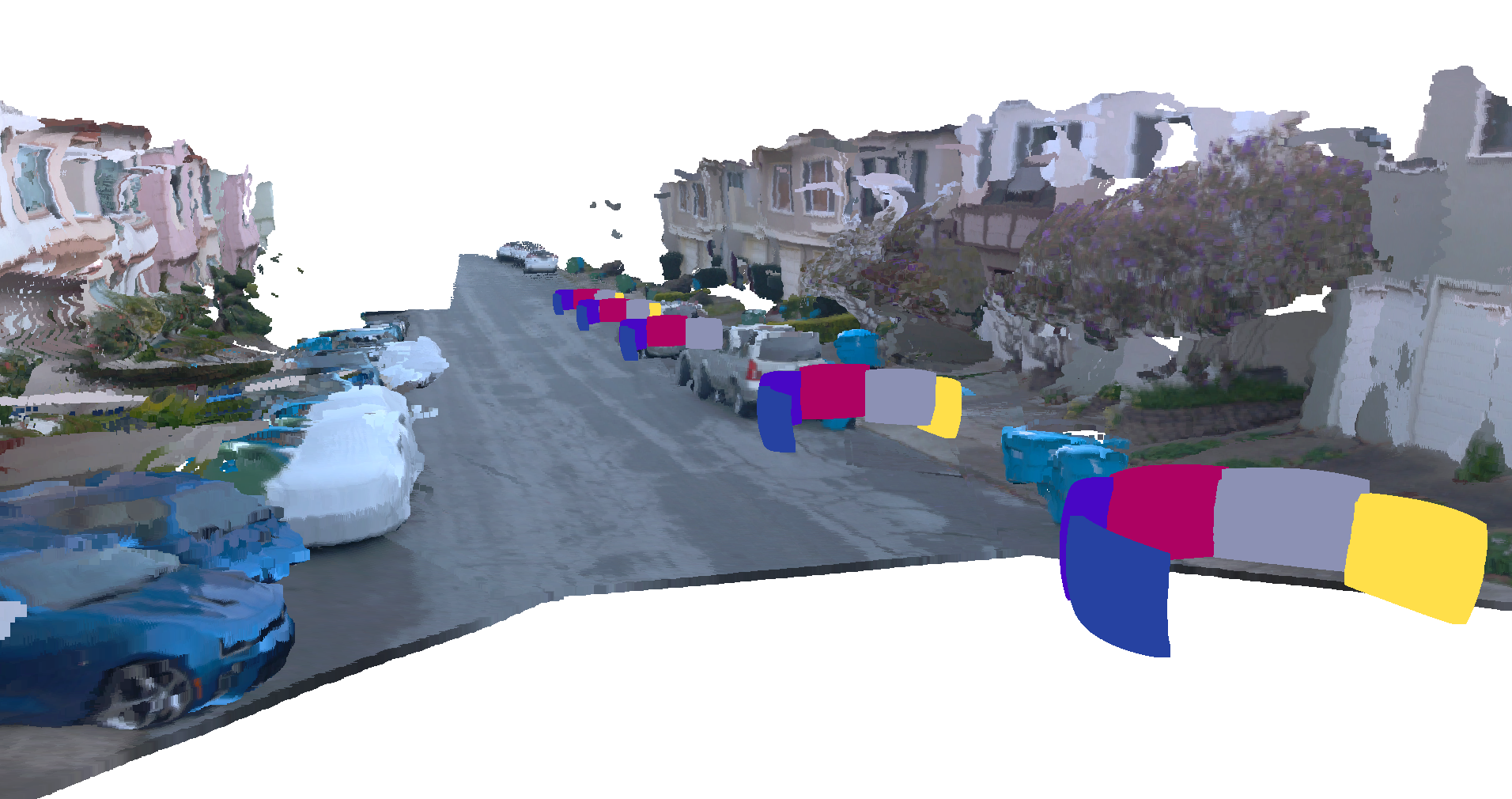}
    \end{minipage}
    \hfill
    \begin{minipage}[b]{0.32\linewidth}
        \centering
        \includegraphics[width=\linewidth]{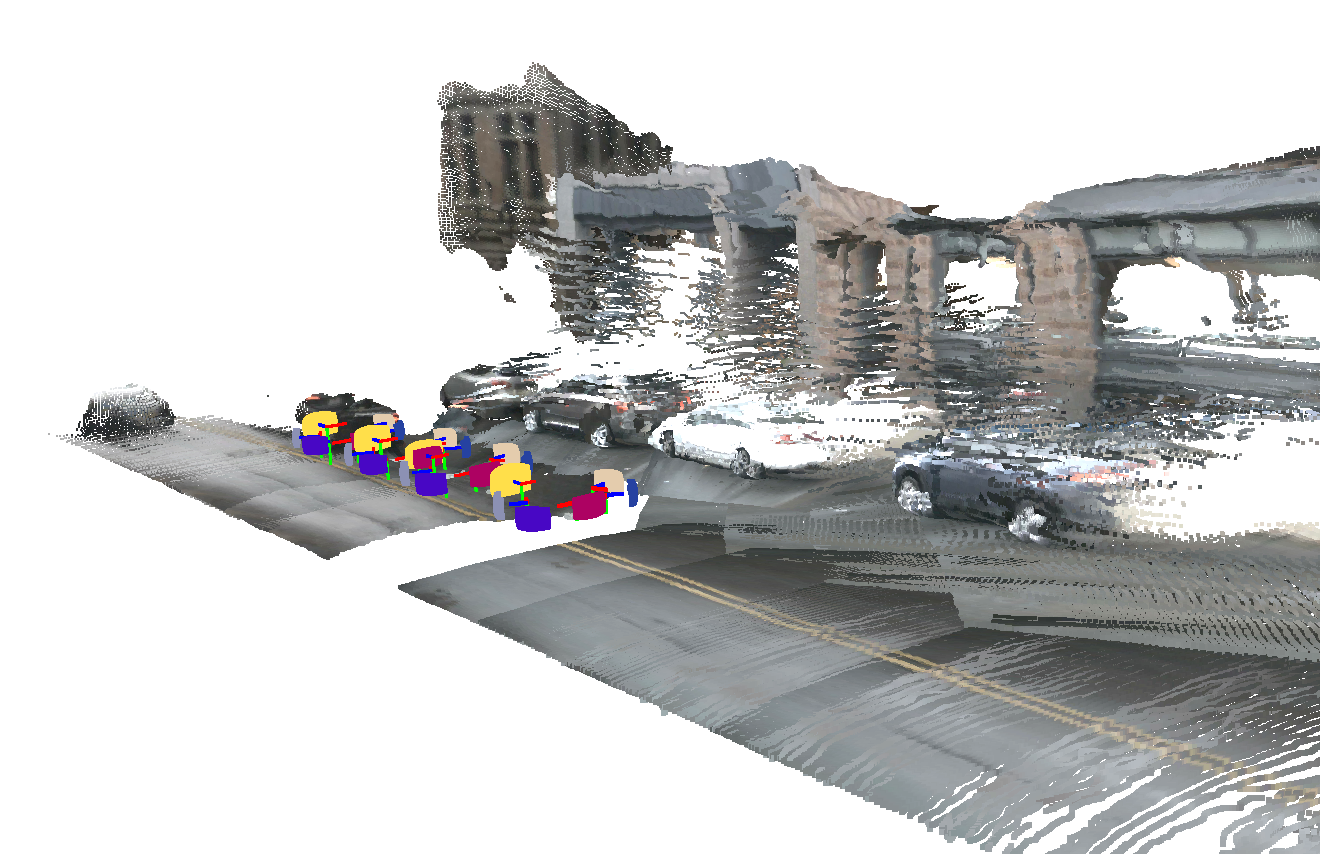}
    \end{minipage}
    \caption{Rig scene reconstructions on Argoverse \cite{wilson2023argoverse2}, Waymo \cite{sun2020waymo}, and nuScenes \cite{nuscenes2019} validation sets.}
    \label{fig:driving_reconstr}
\end{figure}


\section{Additional Visualization}

Figure~\ref{fig:images_embed_ablation} presents qualitative reconstruction results to illustrate the embedding ablations in Section \ref{sec:ablation} of the main paper. While the unstructured model produces visually reasonable reconstructions, some frames are misaligned in position and orientation relative to the rig. Introducing time embeddings helps correct positional drift, while rig pose embeddings improve orientation alignment. The fully calibrated model achieves the highest reconstruction quality and most accurate pose estimates, demonstrating the compounding benefits of both embedding types.


Figure \ref{fig:driving_reconstr} presents additional visualization of Rig3r outputs on diverse rig scenes from Argoverse \cite{wilson2023argoverse2}, Waymo \cite{sun2020waymo}, and nuScenes \cite{nuscenes2019} validation sets. The visualizations include confidence-thresholded pointmaps and global raymaps for rig scenes (color-coded by discovered rig structure), and highlight Rig3R’s consistency under diverse conditions. No post-processing is applied to the 3D points or poses, aside from confidence thresholding and sky masking for visualization.

\section{Estimating Camera Parameters from Raymaps}
\label{sec:pose_from_rmap}

We describe the method used for estimating camera intrinsics and extrinsics from raymaps, as introduced in the Section \ref{sec:raymaps} of the main paper.

\textbf{Intrinsics} At each pixel \( (u, v) \), the ray direction from the raymap is interpreted as a unit vector \( \hat{\mathbf{r}} \) in a global coordinate frame. In the pinhole camera model, the corresponding normalized ray in the camera frame is given by

\begin{equation}
\label{eqn:pixel_to_ray}
\hat{\mathbf{r}}_{\text{cam}} = \frac{1}{\|\cdot\|} \begin{bmatrix} u / f_x \\ v / f_y \\ 1 \end{bmatrix}.
\end{equation}

Here, \( f_x \) and \( f_y \) are focal lengths, and \( (u, v) \) denotes image coordinates relative to a known principal point, which we fix to the image center—a standard simplification in SfM and multiview geometry.

Given two pixels, the angle \( \theta \) between their predicted ray directions must be consistent with the angle computed from the camera model and intrinsics, i.e., $\cos\theta=\hat{\mathbf{r}}^T \hat{\mathbf{r}}'=\hat{\mathbf{r}}_{\text{cam}}^T \hat{\mathbf{r}}_{\text{cam}}'$. Squaring and writing this in terms of camera coordinates gives



\begin{equation}
\label{eqn:iac}
\cos^2 \theta = \frac{(\tilde{\mathbf{u}}^\top \omega \tilde{\mathbf{u}}')^2}{(\tilde{\mathbf{u}}^\top \omega \tilde{\mathbf{u}})(\tilde{\mathbf{u}}'^\top \omega \tilde{\mathbf{u}}')}, \:\: \text{where} \:\: \omega = \mathrm{diag}(1/f_x^2,\; 1/f_y^2,\; 1),
\end{equation}

\noindent $\mathbf{r},\mathbf{r}'$ are a pair of world rays from the raymap and $\tilde{\mathbf{u}},\tilde{\mathbf{u}}'$ the corresponding (homogeneous) image coordinates. This equation constrains the focal lengths and can be solved analytically (simultaneous polynomials) or numerically using multiple pixel pairs. The intrinsic matrix is then formed using the assumed camera center and recovered focals.

As a practical simplification, we can estimate \( f_x \) and \( f_y \) analytically by sampling pixel pairs along the image axes. For example, selecting the optical center and a second pixels at \( (\Delta u, 0) \), we obtain:

\[
f_x = \frac{|\Delta u|}{\tan \theta}, \:\: \text{and similarly for } f_y.
\]

This works well in practice due to the high consistency of Rig3R’s predicted raymaps, which provide stable and geometrically faithful directions across pixel locations and views—enabling accurate and efficient focal length estimation.



\textbf{Extrinsics.} Once intrinsics are estimated, we compute the ray direction \( \hat{\mathbf{r}}_{\text{cam}}^{(i)} \) for each pixel \( (u, v) \) using Equation \ref{eqn:pixel_to_ray}. 
The global raymap predicts the corresponding unit ray directions \( \hat{\mathbf{r}}^{(i)} \) in a shared global reference frame. Since both sets of rays are defined at the same pixel locations, we obtain a dense correspondence between camera-frame and global-frame rays. The relationship between them is a rigid transformation consisting of a single rotation \( \mathbf{R} \), such that


\[
\hat{\mathbf{r}}^{(i)} = \mathbf{R} \hat{\mathbf{r}}_{\text{cam}}^{(i)}.
\]

We solve for the optimal rotation \( \mathbf{R} \) that minimizes angular error across all correspondences using cross-covariance alignment and singular value decomposition (SVD), following~\cite{arun1987least}.











\section{Raymaps vs. Pointmaps for Pose Estimation}
\label{sec:raymap-vs-pointmap}

This section provides further analysis and experimental results comparing raymaps and pointmaps as output representations for pose estimation. While raymaps encode per-pixel ray directions, and can be derived directly from camera intrinsics and extrinsics, pointmaps require predicting full 3D coordinates via per-pixel depth estimation. This makes pointmap-based inference strictly harder and more error-prone—especially in low-texture, reflective, dynamic, or sky regions where depth is ill-posed. Thus we expect pose estimation from raymaps to be more stable than for pointmaps. 


    To test this hypothesis, we evaluate three Rig3R variants: PnP RANSAC \cite{sq-pnp} on the predicted pointmaps with confidence thresholding, the same with sky masking, and pose estimation using closed-form solutions on the global raymap. As shown in Table~\ref{tab:pmap_vs_rmap_pose}, the raymap-based method consistently outperforms both pointmap variants. Even after masking sky pixels—where depth is undefined—pointmap-based estimates remain less accurate and more variable, indicating that relying on intermediate 3D points is suboptimal for pose inference.



\begin{table*}[t]
    \centering
    \renewcommand{\arraystretch}{1}
    \setlength{\tabcolsep}{4pt}
    \begin{tabular}{l|cc|cc|cc}
        \toprule
        \multirow{2}{*}{Method}
        & \multicolumn{2}{c|}{@15° $\uparrow$} 
        & \multicolumn{2}{c|}{@5° $\uparrow$} 
        & \multicolumn{2}{c}{@30° $\uparrow$} \\
        \cmidrule(lr){2-3} \cmidrule(lr){4-5} \cmidrule(lr){6-7}
        & RRA & RTA & RRA & RTA & \multicolumn{2}{c}{mAA} \\
        \midrule
        Rig3R\textsubscript{Calib} (pointmap)            & 25.2 & 26.3 & 1.3 & 4.3 & \multicolumn{2}{c}{7.7} \\
        Rig3R\textsubscript{Calib} (pointmap + sky mask) & \textcolor{orange}{69.2} & \textcolor{orange}{46.8} & \textcolor{orange}{59.7} & \textcolor{orange}{25.4} & \multicolumn{2}{c}{\textcolor{orange}{34.0}} \\
        Rig3R\textsubscript{Calib} (raymap)              & \textcolor{cyan}{\textbf{99.4}} & \textcolor{cyan}{\textbf{91.6}} & \textcolor{cyan}{\textbf{67.4}}  & \textcolor{cyan}{\textbf{77.4}} & \multicolumn{2}{c}{\textcolor{cyan}{\textbf{82.1}}} \\
        \bottomrule
    \end{tabular}
    \caption{
        Comparison of Pointmaps vs Raymaps for pose estimation on Waymo.
    }
    \label{tab:pmap_vs_rmap_pose}
\end{table*}




\begin{figure*}[t]
    \centering
    \includegraphics[width=0.7\linewidth]{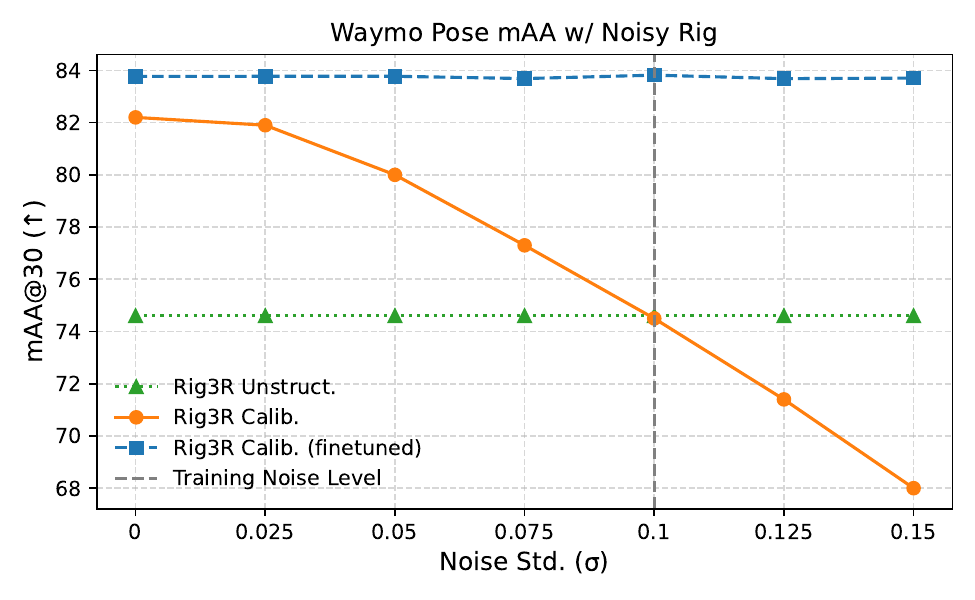}
    \caption{
        Robustness to rig metadata noise before and after finetuning on noisy embeddings. We plot mAA@30 as Gaussian noise is added to rig pose embeddings at inference time. Finetuning with noisy metadata improves performance across all noise levels.
    }
    \label{fig:maa_vs_noise}
\end{figure*}

\section{Sensitivity to Noisy Calibration Embeddings}
\label{sec:finetuning}

We present results of an additional experiment to evaluate Rig3R’s robustness to rig calibration error, which commonly arises in real-world systems due to hardware tolerances, sensor drift, or coarse offline estimation. We additionally test whether training on noisy inputs improves performance when calibration metadata is degraded at inference time.

We simulate noise by independently perturbing rig extrinsics—translation and rotation (roll, pitch, yaw)—with zero-mean Gaussian noise. During training, we use a fixed standard deviation of 0.1. Rig positions are normalized so their average distance from the reference camera is 1, making this noise roughly equivalent to 10 cm deviation for a 1-meter rig. Preliminary tests showed similar performance between Rig3R\textsubscript{Calib} and Rig3R\textsubscript{Unstr} at this noise level, guiding our choice. At inference time, we evaluate robustness across a range of increasingly severe noise levels.

We compare three models: (1) Rig3R without rig embeddings (Unstr.), (2) the original Rig3R trained on clean metadata, and (3) Rig3R finetuned on noisy metadata. As shown in Fig.~\ref{fig:maa_vs_noise}, the clean model performs well under mild noise but degrades steadily as noise increases. In contrast, the finetuned model maintains high performance across all noise levels. The unstructured variant remains flat across noise levels, as it does not use rig embeddings and serves as a lower-bound control. We observe that the base model still stays above the unstructured model even as noise increases, and even has higher performance than the base calibrated model. These results show that training on noisy rig metadata enables Rig3R to remain robust to calibration errors at inference time. 
This enhances robustness to degraded inputs and makes the approach more practical for real-world deployment, where calibration is often approximate but still informative.

\bibliographystyle{unsrtnat}
\bibliography{main}

\clearpage

\end{document}